\documentclass[11pt,a4paper]{article}

\usepackage[T1]{fontenc}
\usepackage[utf8]{inputenc}
\usepackage{lmodern}

\usepackage[margin=2.5cm]{geometry}

\usepackage{amsmath,amssymb}

\usepackage{booktabs}
\usepackage{array}
\usepackage{tabularx}
\usepackage{colortbl}

\usepackage{tikz}
\usetikzlibrary{arrows.meta, positioning, shapes.geometric, calc, fit}

\usepackage{xcolor}
\definecolor{axisblue}{HTML}{2B6CB0}
\definecolor{axisteal}{HTML}{2C7A7B}
\definecolor{cellpop}{HTML}{E6F0FA}
\definecolor{cellempty}{HTML}{F7F7F7}
\definecolor{accent}{HTML}{C05621}
\definecolor{headerblue}{HTML}{EBF4FF}
\definecolor{lightgray}{HTML}{F0F0F0}

\usepackage[colorlinks=true,linkcolor=axisblue,citecolor=axisteal,urlcolor=accent]{hyperref}

\usepackage[numbers,sort&compress]{natbib}

\usepackage{enumitem}
\setlist{nosep,leftmargin=1.5em}
\usepackage{caption}
\title{\textbf{A Two-Dimensional Framework for AI Agent Design Patterns:\\[4pt]
Cognitive Function $\times$ Execution Topology}}

\author{%
  \textbf{Jia Huang}\textsuperscript{1}\quad
  \textbf{Joey Tianyi Zhou}\textsuperscript{1,2}\\[6pt]
  \textsuperscript{1}Agency for Science, Technology and Research (A*STAR), Singapore\\
  \textsuperscript{2}Centre for Frontier AI Research (CFAR), A*STAR\\[4pt]
  \texttt{huang\_jia@a-star.edu.sg}\quad
  \texttt{joey\_zhou@a-star.edu.sg}
}
\date{March 2026 (v1); May 2026 (v2)}

\begin{document}
\maketitle

\begin{abstract}
\noindent Existing frameworks for LLM-based agent architectures describe systems from a single perspective: industry guides (Anthropic, Google, LangChain) focus on \emph{execution topology}---how data flows---while cognitive science surveys focus on \emph{cognitive function}---what the agent does. Neither axis alone disambiguates architecturally distinct systems: the same Orchestrator-Workers topology can implement Plan-and-Execute, Hierarchical Delegation, or Adversarial Verification---three patterns with fundamentally different failure modes and design trade-offs.

We propose a two-dimensional classification that combines (1)~a \textbf{Cognitive Function axis} with seven categories (Perception, Memory, Reasoning, Action, Reflection, Collaboration, Governance) and (2)~an \textbf{Execution Topology axis} with six structural archetypes (Chain, Route, Parallel, Orchestrate, Loop, Hierarchy). The resulting $7 \times 6$ matrix identifies 28 named patterns, 15 with original names. We demonstrate orthogonality through systematic cross-axis analysis, define eight representative patterns in detail, and validate descriptive coverage across four real-world domains (financial lending, legal due diligence, network operations, healthcare triage). Cross-domain analysis yields five empirical laws of pattern selection governing the relationship between environmental constraints (time pressure, action authority, failure cost asymmetry, volume) and architectural choices. The framework provides a principled, framework-neutral, and model-agnostic vocabulary for AI agent architecture design.

\medskip
\noindent\textbf{Keywords:} AI agents, design patterns, taxonomy, cognitive function, execution topology, software architecture, multi-agent systems
\end{abstract}

\section{Introduction}

The rapid deployment of LLM-based agent systems has produced a fragmented landscape of architectural guidance. Every major AI organization has published its own framework for understanding agent architectures:

\begin{itemize}
  \item Anthropic's ``Building Effective Agents''~\cite{anthropic2024agents} identifies six execution topologies (prompt chaining, routing, parallelization, orchestrator-workers, evaluator-optimizer, autonomous agents).
  \item Google's Agent Development Kit~\cite{google2025adk} describes eight workflow patterns organized around sequential, parallel, and loop structures.
  \item LangChain's multi-agent guide~\cite{langchain2025multi} presents four coordination patterns (supervisor, hierarchical, network, handoff).
  \item Andrew Ng's agentic design patterns~\cite{ng2024agentic} identifies four cognitive capabilities (reflection, tool use, planning, multi-agent collaboration).
\end{itemize}

\textbf{The critical observation:} all existing frameworks describe agent architectures from \emph{only one axis}. Industry sources focus on execution topology---\emph{how} data flows. Cognitive surveys~\cite{wang2024survey,xi2023rise,sumers2024cognitive} focus on functional capability---\emph{what} the agent does. Neither alone disambiguates architecturally distinct systems.

Consider the Orchestrator-Workers topology. The same structural wiring diagram serves at least three fundamentally different patterns:

\begin{enumerate}
  \item \textbf{Plan-and-Execute} (Action): a planner decomposes a task into subtasks and dispatches them to executor agents.
  \item \textbf{Hierarchical Delegation} (Collaboration): a manager obtains specialized expertise from domain-specific sub-agents.
  \item \textbf{Observability Harness} (Governance): a central monitor orchestrates logging, tracing, and alerting across agent modules.
\end{enumerate}

These are architecturally distinct systems with different failure modes, different scaling properties, and different testing strategies---yet they share a topology. Without the cognitive function axis, they are indistinguishable. Similarly, a single cognitive function can be realized by multiple topologies: Reasoning (C3) can be implemented as Chain-of-Thought (Chain), Complexity-Based Routing (Route), Parallel Exploration (Parallel), or Iterative Hypothesis Testing (Loop). The choice of topology determines latency, cost, and failure characteristics.

This paper makes four contributions:

\begin{enumerate}
  \item A \textbf{two-dimensional classification framework} that combines cognitive function and execution topology into a single coordinate system (Section~\ref{sec:framework}).
  \item \textbf{Detailed definitions} of eight representative patterns---one per cognitive function plus a governance pattern---sufficient for independent understanding (Section~\ref{sec:patterns}).
  \item A \textbf{systematic orthogonality demonstration} showing that neither axis reduces to the other (Section~\ref{sec:orthogonality}).
  \item A \textbf{coverage evaluation} across four real-world domains that validates the framework's descriptive power and yields five empirical laws of pattern selection (Section~\ref{sec:evaluation}).
\end{enumerate}

\section{The Two-Dimensional Framework}
\label{sec:framework}

\subsection{Design Principles}

Our framework is guided by three principles:

\textbf{Orthogonality.} The two axes must be independently variable. A change in cognitive function should not necessitate a change in execution topology, and vice versa.

\textbf{Completeness.} The axes should cover all capabilities required for production agent systems and all structural archetypes sufficient to compose any agent workflow.

\textbf{Durability.} Categories describe \emph{structural needs} and \emph{structural forms} that persist across framework and model changes. ``Perception'' remains relevant whether the context window is 4K or 2M tokens; ``Loop'' remains relevant whether the loop body is a GPT-4 call or a Claude call.

\subsection{Axis 1: Cognitive Function (What)}

We identify seven cognitive function categories. These are grounded in the cognitive science literature on language agents~\cite{sumers2024cognitive} (Perception, Memory, Reasoning, Action) and extended with three first-class categories (Reflection, Collaboration, Governance) that emerged from production deployment analysis:

\begin{table}[!h]
\centering
\caption{Seven cognitive function categories.}
\label{tab:cognitive}
\small
\begin{tabularx}{\textwidth}{@{}c l X@{}}
\toprule
\textbf{ID} & \textbf{Category} & \textbf{Core Question} \\
\midrule
\rowcolor{lightgray}
C1 & Perception & What information enters the agent's working memory? \\
C2 & Memory & How does the agent store, retrieve, and update knowledge? \\
\rowcolor{lightgray}
C3 & Reasoning & How does the agent deliberate and decide? \\
C4 & Action & How does the agent act on the world through tools? \\
\rowcolor{lightgray}
C5 & Reflection & How does the agent evaluate and improve its own outputs? \\
C6 & Collaboration & How do multiple agents coordinate to solve a problem? \\
\rowcolor{lightgray}
C7 & Governance & How is the agent bounded, observed, and controlled? \\
\bottomrule
\end{tabularx}
\end{table}

The seven categories form a cognitive processing pipeline: an agent perceives input (C1), retrieves relevant knowledge (C2), reasons about what to do (C3), executes actions (C4), evaluates its outputs (C5), coordinates with other agents when needed (C6), and operates within governance constraints throughout (C7). This pipeline is not strictly sequential---agents cycle through Perception-Reasoning-Action loops repeatedly---but the categories are functionally distinct.

\subsection{Axis 2: Execution Topology (How)}

We identify six execution topology archetypes. These subsume the topologies described in existing industry frameworks~\cite{anthropic2024agents,google2025adk}:

\begin{table}[!h]
\centering
\caption{Six execution topology archetypes.}
\label{tab:topology}
\small
\begin{tabularx}{\textwidth}{@{}c l X@{}}
\toprule
\textbf{ID} & \textbf{Archetype} & \textbf{Structure} \\
\midrule
\rowcolor{lightgray}
T1 & Chain & Linear sequential pipeline; output of step $n$ feeds step $n+1$ \\
T2 & Route & Conditional branching; a classifier dispatches to specialized handlers \\
\rowcolor{lightgray}
T3 & Parallel & Concurrent fan-out with aggregation; independent subtasks run simultaneously \\
T4 & Orchestrate & Central coordinator delegates to workers and synthesizes results \\
\rowcolor{lightgray}
T5 & Loop & Iterative refinement with explicit exit conditions \\
T6 & Hierarchy & Nested multi-level delegation; each level can use any other topology \\
\bottomrule
\end{tabularx}
\end{table}

\subsection{The $7\times6$ Pattern Matrix}

The Cartesian product yields a $7 \times 6 = 42$ cell matrix. We identify 28 named patterns occupying these cells; the remaining 14 cells are either structurally redundant or not yet observed in practice. Table~\ref{tab:matrix} shows the complete matrix. Cells marked with $\bigstar$ carry original names coined in this work.

\begin{table}[!t]
\centering
\caption{The $7\times6$ pattern matrix. 28 named patterns; $\bigstar$ = original name coined in this work.}
\label{tab:matrix}
\small
\setlength{\tabcolsep}{3pt}
\begin{tabularx}{\textwidth}{@{}l|X X X X X X@{}}
\toprule
& \textbf{T1 Chain} & \textbf{T2 Route} & \textbf{T3 Parallel} & \textbf{T4 Orchestrate} & \textbf{T5 Loop} & \textbf{T6 Hierarchy} \\
\midrule
\rowcolor{lightgray}
\textbf{C1} & Semantic Compact.$\bigstar$ & Context Triage$\bigstar$ & Multi-Modal Fusion & Progressive Disc.$\bigstar$ & --- & --- \\
\textbf{C2} & RAG Pipeline & Hierarchical Ret.$\bigstar$ & --- & Progress Track.$\bigstar$ & Failure Journal$\bigstar$ & --- \\
\rowcolor{lightgray}
\textbf{C3} & Chain-of-Thought & Complexity Rte.$\bigstar$ & Parallel Explor. & --- & Iterative Hyp.$\bigstar$ & --- \\
\textbf{C4} & Prompt Chaining & Tool Dispatch & --- & Plan-and-Execute & --- & Guardrail Sand.$\bigstar$ \\
\rowcolor{lightgray}
\textbf{C5} & Generator-Critic & Skill Package$\bigstar$ & --- & --- & Self-Heal Loop$\bigstar$ & Exp.\ Replay \\
\textbf{C6} & Handoff Chain & --- & Fan-Out/Gather & --- & Adversarial Rev. & Hier.\ Deleg. \\
\rowcolor{lightgray}
\textbf{C7} & --- & Approval Gate$\bigstar$ & Prog.\ Commit.$\bigstar$ & Observ.\ Harness$\bigstar$ & --- & Blast Radius$\bigstar$ \\
\bottomrule
\end{tabularx}
\end{table}

\section{Representative Pattern Definitions}
\label{sec:patterns}

We define eight representative patterns---at least one per cognitive function row---in sufficient detail for independent understanding. Each definition follows the format: \emph{coordinates, problem, architectural solution, and engineering trade-offs}. The remaining 19 patterns follow the same template.

\subsection{Context Triage (C1 $\times$ T2: Perception $\times$ Route)}

\textbf{Problem.} An agent at task start has access to many information sources: the user's message, conversation history, project files, documentation, tool outputs, retrieved knowledge, and environmental metadata. The context window cannot hold everything. The naive approach---first-in-first-out, truncate when full---fails because relevance is not correlated with recency.

\textbf{Solution.} Context Triage applies emergency-room triage logic to information selection. Every information source is classified by priority (P0--P3), and a routing function dispatches each source to the appropriate treatment: P0 (always load), P1 (load if relevant), P2 (load on demand), P3 (never load). The routing function evaluates source relevance against the current task description, token budget constraints, and cache-friendliness. Claude Code's five-level CLAUDE.md hierarchy (Enterprise $\to$ User $\to$ Project $\to$ Rules $\to$ Local) is a production implementation of this pattern.

\textbf{Trade-offs.} Higher triage accuracy reduces context noise but increases routing latency. Over-aggressive filtering risks starving the agent of critical information; under-filtering dilutes attention quality~\cite{liu2024lost}.

\subsection{RAG Pipeline (C2 $\times$ T1: Memory $\times$ Chain)}

\textbf{Problem.} The agent needs knowledge beyond what fits in its context window or its training data. The knowledge may be domain-specific, frequently updated, or proprietary.

\textbf{Solution.} Retrieval-Augmented Generation~\cite{lewis2020rag} implements an ``open-book exam'': query $\to$ retrieve $\to$ rerank $\to$ generate. The chain topology ensures each step's output feeds the next. The retrieval step converts the agent's information need into a vector query against an external knowledge store; the reranking step filters and orders results by relevance; the generation step synthesizes an answer conditioned on retrieved evidence. MemGPT~\cite{packer2023memgpt} extends this with virtual context management---paging data between the context window and external storage using OS-inspired memory management.

\textbf{Trade-offs.} Each retrieved chunk consumes token budget. Retrieving 10 chunks at 500 tokens each costs 5,000 tokens that cannot be spent on reasoning. The architect must balance recall (more chunks, more evidence) against precision (fewer chunks, more attention per chunk).

\subsection{Complexity-Based Routing (C3 $\times$ T2: Reasoning $\times$ Route)}

\textbf{Problem.} Agent workloads contain queries of widely varying difficulty. Applying deep Chain-of-Thought reasoning with 64K thinking tokens to ``What are your store hours?'' wastes compute; applying a quick heuristic to a multi-step diagnostic task produces errors.

\textbf{Solution.} A lightweight classifier evaluates each incoming query and routes it to the appropriate reasoning depth: System~1 (direct response, $\sim$500 tokens) for simple queries, System~2 (Chain-of-Thought, $\sim$8K tokens) for moderate queries, and extended deliberation ($\sim$64K tokens) for complex queries. This mirrors Kahneman's dual-process theory~\cite{kahneman2011thinking}: the architecture decides \emph{how deeply} to think before thinking. RouteLLM~\cite{ong2024routellm} demonstrated 85\% cost reduction with minimal quality loss by routing between strong and weak models.

\textbf{Trade-offs.} Classifier accuracy determines system performance. Misrouting a complex query to System~1 produces errors; misrouting a simple query to System~2 wastes tokens. At 100,000 daily queries, the difference between \$0.0015 and \$0.19 per query is \$18,850/day.

\subsection{Plan-and-Execute (C4 $\times$ T4: Action $\times$ Orchestrate)}

\textbf{Problem.} A complex task requires multiple tool calls where the execution order depends on intermediate results. A single-step approach cannot decompose the task; a purely sequential chain is too rigid to handle dynamic dependencies.

\textbf{Solution.} Separate strategy from tactics: a \emph{planner} agent decomposes the task into a directed acyclic graph (DAG) of subtasks, and an \emph{executor} agent (or pool of executors) carries out each subtask. The planner can use a cheaper model; the executors handle tool invocation with domain-specific prompts. This is the Saga pattern from distributed systems~\cite{garcia1987sagas} adapted for agent workflows: each subtask is a compensatable action, and the planner manages the overall transaction.

\textbf{Trade-offs.} Plan quality determines execution efficiency. Over-decomposition creates unnecessary coordination overhead; under-decomposition creates subtasks too complex for executors. The separation also introduces latency---planning before executing is slower than immediate execution for simple tasks.

\subsection{Generator-Critic (C5 $\times$ T1: Reflection $\times$ Chain)}

\textbf{Problem.} An agent's first-draft output is functional but imperfect. The agent needs to improve this output through structured critique before delivery, but the refinement should terminate quickly---a bounded short chain rather than an open-ended loop.

\textbf{Solution.} Separate generation from evaluation and chain them: generate $\to$ critique $\to$ revise. The critical design choice is the \emph{feedback source} and the stopping condition. Huang et al.~\cite{huang2024selfcorrect} (ICLR 2024) demonstrated that LLMs cannot reliably self-correct without external feedback. Three variants address this: (1)~self-critique using different prompts for generation and evaluation, (2)~cross-model critique where a separate model evaluates, and (3)~tool-grounded critique where a test suite, linter, or calculator provides deterministic feedback. CRITIC~\cite{gou2024critic} showed that tool-interactive critiquing consistently outperforms pure self-critique. Self-Refine~\cite{madaan2023selfrefine} demonstrated $\sim$20\% absolute improvement across seven tasks through 2--4 iterations. In practice, production deployments converge in 1--2 critique passes, which makes Chain (bounded short pipeline with quality-threshold exit) a more accurate placement than Loop. The Loop topology is reserved for the related but distinct Self-Heal pattern, in which iteration continues until an external verifier (test suite, schema validator, or runtime check) passes.

\textbf{Trade-offs.} Each iteration costs tokens. Convergence requires a stopping criterion---a quality threshold or iteration budget---to prevent over-editing or oscillation. Generator-Critic and Self-Heal Loop differ structurally by stopping condition: Generator-Critic exits on subjective quality judgment (chain-like), Self-Heal exits on deterministic verification (loop-like).

\subsection{Fan-Out/Gather (C6 $\times$ T3: Collaboration $\times$ Parallel)}

\textbf{Problem.} A task is decomposable into independent subtasks that can be processed simultaneously. A single agent processing them sequentially would take $n\times$ the time.

\textbf{Solution.} A coordinator fans out subtasks to $n$ worker agents running in parallel, then gathers and aggregates their results. Each worker operates in an isolated context window containing only its subtask. The coordinator handles result synthesis, conflict resolution, and quality assessment. Du et al.~\cite{du2024multi} showed that multiagent debate among multiple LLM instances improves factuality and reasoning, with consensus-based aggregation outperforming single-agent baselines on arithmetic and strategic reasoning benchmarks. However, naive aggregation without structured debate protocols can amplify errors when agents produce conflicting outputs.

\textbf{Trade-offs.} $n$ workers consume $n\times$ tokens. The aggregation step is the quality bottleneck---naive concatenation of worker outputs produces incoherent results. Worker independence must be genuine; interdependent subtasks force sequential execution regardless of topology.

\subsection{Approval Gate (C7 $\times$ T2: Governance $\times$ Route)}

\textbf{Problem.} An agent that acts on the real world faces a dilemma: too many approval prompts cause ``approval fatigue'' (users click approve reflexively), while too few allow irreversible damage.

\textbf{Solution.} Route every agent action through a three-stage evaluation: (1)~\textsc{Deny} rules (absolute priority---block dangerous actions unconditionally), (2)~\textsc{Allow} rules (auto-approve low-risk actions to reduce noise), (3)~\textsc{Human} gate (residual---anything not denied or allowed reaches the human). Actions are classified along two dimensions: reversibility (can the action be undone?) and impact (how much damage if wrong?). Claude Code implements this as a five-tier permission system ranging from \texttt{default} (prompt for everything dangerous) to \texttt{bypassPermissions} (no gates).

\textbf{Trade-offs.} Classifier accuracy is critical. Under-classification blocks safe actions and frustrates users; over-classification allows dangerous actions through. The key design variable is the granularity of the reversibility/impact classification: too coarse and safe actions are blocked; too fine and the classifier itself becomes a maintenance burden.

\subsection{Blast Radius Control (C7 $\times$ T6: Governance $\times$ Hierarchy)}

\textbf{Problem.} Even with approval gates, an agent may cause damage through unexpected tool interactions, cascading failures, or actions that are individually safe but collectively dangerous.

\textbf{Solution.} Nested containment hierarchies limit the maximum damage any single action can cause. Each level constrains the child: process sandbox $\to$ filesystem isolation $\to$ network restrictions $\to$ API rate limits $\to$ budget caps. The hierarchy topology is essential: each containment layer can enforce different policies, and the outermost layer represents the organizational risk boundary. Codex CLI's sandbox enables a ``full-auto'' mode precisely because the sandbox guarantees bounded damage.

\textbf{Trade-offs.} Tighter containment reduces blast radius but also reduces agent capability. Finding the minimum viable containment---the tightest sandbox that still permits the agent to accomplish its task---is the governance architect's central challenge.

\section{Orthogonality Demonstration}
\label{sec:orthogonality}

To validate that the two axes are genuinely independent, we show that (a)~a single topology serves multiple cognitive functions, and (b)~a single cognitive function is served by multiple topologies.

\subsection{Same Topology, Different Cognitive Functions}

The \textbf{Loop} topology (T5) serves at least three distinct cognitive purposes within the matrix:

\begin{itemize}
  \item \textbf{Failure Journal} (C2): iteratively recording and consolidating error patterns from past executions.
  \item \textbf{Iterative Hypothesis Testing} (C3): alternating hypothesis generation with evidence gathering through environment interaction.
  \item \textbf{Self-Heal Loop} (C5): iteratively diagnosing and repairing failures until an external verifier (test suite, schema check, or runtime probe) passes.
\end{itemize}

All three share the same \texttt{while(!done)} control structure. They differ entirely in \emph{what cognitive function} the loop serves: memory consolidation, hypothesis testing, or deterministic-verify repair. A fourth use of Loop---the reason-act interleaving popularized by ReAct~\cite{yao2023react}---is recognized as the agent's outer execution loop (the framework preamble within which all tactical patterns operate) rather than as a discrete tactical pattern occupying a matrix cell. We discuss this scope distinction in Section~\ref{sec:scope}.

\subsection{Same Cognitive Function, Different Topologies}

\textbf{Reasoning} (C3) can be implemented via at least four topologies:

\begin{itemize}
  \item \textbf{Chain-of-Thought} (T1 Chain): linear step-by-step decomposition~\cite{wei2022chain}.
  \item \textbf{Complexity-Based Routing} (T2 Route): dispatching queries to different reasoning depths.
  \item \textbf{Parallel Exploration} (T3 Parallel): tree or graph search across multiple reasoning branches simultaneously.
  \item \textbf{Iterative Hypothesis Testing} (T5 Loop): probe-observe-adjust cycles where the agent reasons \emph{through} environment interaction.
\end{itemize}

The choice of topology determines latency (Chain is fastest, Loop is slowest), cost (Parallel is most expensive), and completeness (Chain may miss alternatives that Parallel would find). This confirms that neither axis reduces to the other: knowing the cognitive function does not determine the topology, and vice versa.

\section{Comparison with Existing Frameworks}

\begin{table}[!h]
\centering
\caption{Comparison with existing agent architecture resources.}
\label{tab:comparison}
\small
\setlength{\tabcolsep}{4pt}
\begin{tabularx}{\textwidth}{@{}l c c c c@{}}
\toprule
\textbf{Resource} & \textbf{Cognitive Axis?} & \textbf{Topology Axis?} & \textbf{\# Patterns} & \textbf{Framework-Neutral?} \\
\midrule
\rowcolor{lightgray}
Anthropic~\cite{anthropic2024agents}   & No  & Yes (6) & 6  & Yes \\
Google ADK~\cite{google2025adk}         & No  & Yes (8) & 8  & No  \\
\rowcolor{lightgray}
LangChain~\cite{langchain2025multi}     & No  & Yes (4) & 4  & No  \\
Andrew Ng~\cite{ng2024agentic}          & Yes (4) & No  & 4  & Yes \\
\rowcolor{lightgray}
Wang et al.~\cite{wang2024survey}       & Yes (4) & No  & --- & Yes \\
Sumers et al.~\cite{sumers2024cognitive} & Yes (4) & No & --- & Yes \\
\rowcolor{lightgray}
Liu et al.~\cite{liu2024agent}          & No  & No  & 18 & Yes \\
Dao et al.~\cite{dao2026agentic}        & Yes (5) & No  & 12 & Yes \\
\midrule
\rowcolor{headerblue}
\textbf{This work} & \textbf{Yes (7)} & \textbf{Yes (6)} & \textbf{28} & \textbf{Yes} \\
\bottomrule
\end{tabularx}
\end{table}

No existing framework provides both axes simultaneously. Liu et al.~\cite{liu2024agent} catalog 18 agent design patterns but organize them by a flat category system without orthogonal axes, making it impossible to distinguish patterns that share a topology but serve different cognitive functions. Dao et al.~\cite{dao2026agentic} use a system-theoretic lens with five functional categories but do not cross them with execution topologies. Our contribution is not the individual axes---both have precedents---but their \emph{systematic combination} into a single coordinate system that enables unambiguous pattern identification.

\section{Coverage Evaluation}
\label{sec:evaluation}

To validate the framework's descriptive power, we apply a six-step Pattern Selection Methodology (Bound $\to$ Map $\to$ Topology $\to$ Select $\to$ Impact $\to$ Build) to four real-world domains. Each domain was chosen to stress different aspects of the framework: different time constraints, different volumes, different risk profiles, and different governance requirements.

\subsection{Four Domain Case Studies}

\begin{table}[!h]
\centering
\caption{Four case study domains with structurally different architectures derived from the same pattern catalog.}
\label{tab:casestudies}
\small
\begin{tabularx}{\textwidth}{@{}l X X X X@{}}
\toprule
& \textbf{Financial Lending} & \textbf{Legal Due Diligence} & \textbf{Network Operations} & \textbf{Healthcare Triage} \\
\midrule
\rowcolor{lightgray}
\textbf{Task} & SME loan assessment & M\&A contract review & NOC alert handling & ED patient triage \\
\textbf{Time budget} & 4 hours & 8 hours & 5 minutes & 60 seconds \\
\rowcolor{lightgray}
\textbf{Volume} & 1 case & 500 contracts & Continuous stream & 1 patient \\
\textbf{Primary topology} & Orchestrate & Hierarchy & Route & Chain \\
\rowcolor{lightgray}
\textbf{Patterns selected} & 7 & 8 & 9 & 7 \\
\textbf{Action authority} & Recommend only & Recommend only & Auto (P3/P4) & Recommend only \\
\rowcolor{lightgray}
\textbf{Governance focus} & Audit trail & Data isolation & Blast radius & Asymmetric safety \\
\bottomrule
\end{tabularx}
\end{table}

\textbf{Financial Lending.} An agent assisting bank credit officers with SME loan assessment. Seven patterns selected: Context Triage (filter applicant documents by relevance), RAG Pipeline (retrieve regulatory rules and industry benchmarks), Complexity-Based Routing (route simple applications to fast-track, complex ones to deep analysis), Iterative Hypothesis Testing (probe financial statements for inconsistencies), Generator-Critic (regulatory compliance review), Approval Gate (human officer makes final decision), and Observability Harness (audit logging for regulatory compliance). The 4-hour budget permits the Orchestrate topology with deep per-pattern execution.

\textbf{Legal Due Diligence.} An agent reviewing 500 contracts for M\&A due diligence. Eight patterns selected, notably adding Fan-Out/Gather (parallel contract processing) and Hierarchical Delegation (partner $\to$ associate $\to$ clause-level review) to handle volume. The Hierarchy topology is driven by Law~4 (volume determines collaboration needs).

\textbf{Network Operations.} An agent handling telecom NOC alerts within a 5-minute SLA. Nine patterns selected, with Route as primary topology to classify alerts by severity and signature match. Blast Radius Control is critical: the agent can auto-execute remediation for P3/P4 alerts but must escalate P1/P2. The tight time budget limits pattern depth (Law~1).

\textbf{Healthcare Triage.} An agent assisting ED triage nurses with patient acuity assessment. Seven patterns within a 60-second budget force the Chain topology (the simplest and fastest). Generator-Critic is parameterized with extreme asymmetry (Law~3): the critic is biased toward upgrading acuity, because under-triage (sending a critical patient to the waiting room) is catastrophically worse than over-triage.

\subsection{Five Laws of Pattern Selection}

From cross-domain comparison, five invariant principles emerge:

\textbf{Law 1: Time pressure determines architectural complexity.} Days afford Hierarchy + Orchestrate (10+ patterns); hours afford Orchestrate (7--8); minutes afford Route + Loop (5--7); seconds afford Chain only (3--5). \emph{The first fix for a slow prototype is not ``optimize each pattern'' but ``remove a pattern.''}

\textbf{Law 2: Action authority determines governance pattern.} Advisory-only systems need Approval Gate (human decides). Systems with low-risk auto-execution need Blast Radius Control (pre-compute impact). Systems with high-risk irreversible actions need Guardrail Sandwich (pre- and post-checks). Mixed systems need tiered governance.

\textbf{Law 3: Failure cost asymmetry reshapes reflection.} When false positives and false negatives have symmetric costs (lending), the Generator-Critic optimizes for accuracy. When costs are extremely asymmetric (healthcare: under-triage is fatal), the critic is deliberately biased toward the safe error.

\textbf{Law 4: Volume determines collaboration needs.} Single-item processing needs no collaboration patterns. Moderate volume (10--50 items) needs Fan-Out/Gather. High volume (100--500) needs Hierarchical Delegation + Fan-Out/Gather. Continuous streams need Route + auto-scaling.

\textbf{Law 5: Same pattern, different parameterization.} The same pattern (e.g., Generator-Critic) appears in all four domains but behaves differently: 5-minute regulatory review (lending), 30-second sanity check (network), biased safety override (healthcare). \emph{A pattern is a structural template, not a behavioral prescription.} The template provides the HOW; the domain provides the WHAT and WHY.

\subsection{Cross-Pattern Analysis}

Four patterns appear in three or more of the four case studies: Context Triage, RAG Pipeline, Complexity-Based Routing, and Generator-Critic. Their ubiquity suggests these are \emph{foundational} patterns---required by most production agent systems regardless of domain. In contrast, Blast Radius Control and Fan-Out/Gather appear only when specific domain constraints (autonomous action authority, high volume) are present, suggesting they are \emph{conditional} patterns triggered by environmental factors.

The framework's empty cells also carry information. C5 (Reflection) now has four populated cells out of six (Chain, Route, Loop, Hierarchy), leaving Parallel and Orchestrate open---an interesting open territory given the structural fit of those topologies for multi-critic ensembles. We hypothesize that as agent systems mature, patterns such as \emph{Parallel Reflection} (multiple critics evaluating simultaneously) and \emph{Reflection Orchestrate} (a meta-critic dispatching outputs to domain-specific evaluators) will emerge.

\section{Discussion}

\subsection{Scope and Limitations}
\label{sec:scope}

\textbf{Non-perfect orthogonality.} Some combinations are more natural than others. Governance patterns tend toward Route and Hierarchy topologies because gate-keeping and containment are inherently hierarchical. We report the framework honestly: 28 of 42 cells are populated (67\%), and the distribution is not uniform. The empty cells may represent genuinely impossible combinations, under-explored territory, or artifacts of current technology limitations.

\textbf{Pattern granularity.} The 28-pattern count reflects a judgment call about granularity. One could split Generator-Critic into self-critique and cross-model-critique (two patterns) or merge Chain-of-Thought with Prompt Chaining (reducing the count). We chose a granularity that maximizes architectural discrimination while remaining memorable.

\textbf{Tactical patterns vs.\ the outer execution loop.} The matrix captures \emph{tactical} patterns---bounded architectural primitives an engineer composes into an agent system. It deliberately excludes the agent's \emph{outer execution loop}, the recurring reason-act cycle popularized by ReAct~\cite{yao2023react}, because that loop is the runtime substrate within which all tactical patterns operate, not a peer pattern alongside them. Treating the outer loop as a matrix cell would conflate scope: every tactical pattern in the matrix executes \emph{inside} some form of agent loop. We therefore treat the outer ReAct-style loop as framework preamble (Section~2) rather than as a discrete entry. This is the same scope distinction by which the Gang of Four catalog did not list ``the object'' as a 24th design pattern.

\textbf{Revision notes (v2, May 2026).} This version refines the matrix relative to v1 in three respects: (i) Generator-Critic moves from C5$\times$Loop to C5$\times$Chain to reflect its typical bounded 1--2 iteration stopping condition; (ii) Self-Heal Loop is introduced at C5$\times$Loop to disambiguate iterative deterministic-verify repair from chain-style critique refinement; (iii) the ReAct outer execution loop, previously listed at C4$\times$Loop in v1, is reclassified as framework preamble per the scope distinction above. Net pattern count: 28.

\textbf{Temporal validity.} The framework is designed for durability, but the specific patterns will evolve. As reasoning models internalize Chain-of-Thought, the architectural Chain-of-Thought pattern may become less relevant---replaced by budget-aware routing to reasoning models. The framework accommodates this: the pattern evolves within its cell, but the coordinate system remains stable.

\subsection{Positioning in Software Engineering History}

We position agent design patterns as the third generation of a 30-year software engineering tradition:

\begin{enumerate}
  \item \textbf{Object-Oriented Patterns} (1994): Gamma et al.~\cite{gamma1994design} responded to the challenge of object composition in deterministic systems. 23 patterns organized by purpose (creational, structural, behavioral) and scope.
  \item \textbf{Enterprise/Distributed Patterns} (2000s): Fowler~\cite{fowler2002patterns}, Hohpe and Woolf~\cite{hohpe2003enterprise} responded to the challenge of distributed system integration. Patterns organized by architectural layer and communication style.
  \item \textbf{Agent Patterns} (2024--): Our framework responds to the challenge of probabilistic, tool-using, multi-agent systems. Patterns organized by cognitive function and execution topology.
\end{enumerate}

Each generation responded to a fundamental shift in system assumptions. Agent patterns respond to the shift from deterministic to probabilistic execution, from compile-time to runtime tool selection, and from single-process to multi-agent coordination.

\section{Conclusion}

We have presented a two-dimensional framework that classifies AI agent design patterns along cognitive function and execution topology axes. The $7 \times 6$ matrix identifies 28 named patterns and provides a coordinate system for unambiguous pattern identification. Our coverage evaluation across four real-world domains demonstrates that the same pattern catalog produces structurally different architectures when different domain constraints are applied, and yields five empirical laws governing pattern selection.

The framework is designed to be framework-neutral, model-agnostic, and durable. As the underlying models and frameworks evolve, the coordinate system---what the agent needs to do (cognitive function) crossed with how it is structurally organized (execution topology)---remains stable. We invite the community to validate, extend, and challenge this framework through additional case studies and empirical evaluation.

An extended engineering treatment of this framework, including the 28 patterns implemented in Python with a running case study (Argus, a code-review agent), the six-step pattern-selection methodology, and the five empirical laws applied to additional production deployments, appears in \href{https://hubs.la/Q04hCssN0}{\emph{Designing AI Agents: Principles, patterns, and best practices}}~\cite{huang2026agent}.

\bibliographystyle{unsrtnat}

\begin{thebibliography}{24}

\bibitem{anthropic2024agents}
E.~Schluntz and B.~Zhang,
``Building effective agents,''
Anthropic Research Blog, Dec.~2024.

\bibitem{google2025adk}
Google Cloud,
``Agent Development Kit: A flexible framework for building multi-agent systems,''
Google Developers Blog, Apr.~2025.

\bibitem{langchain2025multi}
H.~Chase et al.,
``LangGraph: Multi-agent workflows,''
LangChain Documentation, Feb.~2025.

\bibitem{ng2024agentic}
A.~Ng,
``What's next for AI agentic workflows,''
Sequoia Capital AI Ascent, Mar.~2024.

\bibitem{wang2024survey}
L.~Wang, C.~Ma, X.~Feng, et al.,
``A survey on large language model based autonomous agents,''
\emph{Frontiers of Computer Science}, vol.~18, art.~186345, 2024.

\bibitem{xi2023rise}
Z.~Xi, W.~Chen, X.~Guo, et al.,
``The rise and potential of large language model based agents: A survey,''
arXiv:2309.07864, 2023.

\bibitem{sumers2024cognitive}
T.~R.~Sumers, S.~Yao, K.~Narasimhan, and T.~L.~Griffiths,
``Cognitive architectures for language agents,''
\emph{Foundations and Trends in Machine Learning}, vol.~17, no.~6, pp.~882--971, 2024.

\bibitem{gamma1994design}
E.~Gamma, R.~Helm, R.~Johnson, and J.~Vlissides,
\emph{Design Patterns: Elements of Reusable Object-Oriented Software}.
Addison-Wesley, 1994.

\bibitem{fowler2002patterns}
M.~Fowler,
\emph{Patterns of Enterprise Application Architecture}.
Addison-Wesley, 2002.

\bibitem{liu2024agent}
Y.~Liu, S.~K.~Lo, Q.~Lu, et al.,
``Agent design pattern catalogue: A collection of architectural patterns for foundation model based agents,''
\emph{J.\ Systems and Software}, vol.~220, art.~112278, 2024.

\bibitem{dao2026agentic}
M.-D.~Dao, Q.~M.~Le, H.~T.~Lam, et al.,
``Agentic design patterns: A system-theoretic framework,''
arXiv:2601.19752, 2026.

\bibitem{liu2024lost}
N.~F.~Liu, K.~Lin, J.~Hewitt, et al.,
``Lost in the middle: How language models use long contexts,''
\emph{Transactions of the Association for Computational Linguistics}, vol.~12, pp.~157--173, 2024.

\bibitem{lewis2020rag}
P.~Lewis, E.~Perez, A.~Piktus, et al.,
``Retrieval-augmented generation for knowledge-intensive NLP tasks,''
\emph{Advances in Neural Information Processing Systems}, vol.~33, pp.~9459--9474, 2020.

\bibitem{packer2023memgpt}
C.~Packer, S.~Wooders, K.~Lin, et al.,
``MemGPT: Towards LLMs as operating systems,''
arXiv:2310.08560, 2023.

\bibitem{kahneman2011thinking}
D.~Kahneman,
\emph{Thinking, Fast and Slow}.
Farrar, Straus and Giroux, 2011.

\bibitem{ong2024routellm}
I.~Ong, A.~Almahairi, V.~Wu, et al.,
``RouteLLM: Learning to route LLMs with preference data,''
arXiv:2406.18665, 2024.

\bibitem{garcia1987sagas}
H.~Garcia-Molina and K.~Salem,
``Sagas,''
\emph{ACM SIGMOD Record}, vol.~16, no.~3, pp.~249--259, 1987.

\bibitem{huang2024selfcorrect}
J.~Huang, X.~Chen, S.~Mishra, et al.,
``Large language models cannot self-correct reasoning yet,''
\emph{Proc.\ ICLR}, 2024.

\bibitem{gou2024critic}
Z.~Gou, Z.~Shao, Y.~Gong, et al.,
``CRITIC: Large language models can self-correct with tool-interactive critiquing,''
\emph{Proc.\ ICLR}, 2024.

\bibitem{madaan2023selfrefine}
A.~Madaan, N.~Tandon, P.~Gupta, et al.,
``Self-Refine: Iterative refinement with self-feedback,''
\emph{Advances in Neural Information Processing Systems}, vol.~36, 2023.

\bibitem{du2024multi}
Y.~Du, S.~Li, A.~Torralba, J.~B.~Tenenbaum, and I.~Mordatch,
``Improving factuality and reasoning in language models through multiagent debate,''
\emph{Proc.\ ICML}, 2024.
arXiv:2305.14325.

\bibitem{yao2023react}
S.~Yao, J.~Zhao, D.~Yu, et al.,
``ReAct: Synergizing reasoning and acting in language models,''
\emph{Proc.\ ICLR}, 2023.

\bibitem{wei2022chain}
J.~Wei, X.~Wang, D.~Schuurmans, et al.,
``Chain-of-thought prompting elicits reasoning in large language models,''
\emph{Advances in Neural Information Processing Systems}, vol.~35, pp.~24824--24837, 2022.

\bibitem{hohpe2003enterprise}
G.~Hohpe and B.~Woolf,
\emph{Enterprise Integration Patterns: Designing, Building, and Deploying Messaging Solutions}.
Addison-Wesley, 2003.

\bibitem{huang2026agent}
J.~Huang,
\href{https://hubs.la/Q04hCssN0}{\emph{Designing AI Agents: Principles, patterns, and best practices}}.
Manning Publications, 2026.

\end{thebibliography}

\end{document}